\documentclass{article}
\usepackage{amssymb}
\usepackage{hyperref}
\usepackage{amsmath}
\usepackage{xcolor}

\usepackage[final]{corl_2025} %
\usepackage{booktabs}
\usepackage{caption}
\usepackage{wrapfig}
\usepackage{graphicx}
\usepackage[export]{adjustbox}
\usepackage{graphicx}
\usepackage{xspace}

\title{

Eye, Robot: Learning to Look to Act \\with a BC-RL Perception-Action Loop

}

\author{
\normalfont
Justin Kerr\quad
Kush Hari\quad
Ethan Weber\quad
Chung Min Kim\quad
Brent Yi\\
Tyler Bonnen\quad
Ken Goldberg\quad
Angjoo Kanazawa
  \\
  \vspace{-0.7em}
  \\
  \url{https://www.eyerobot.net/}\\
  \vspace{-0.7em}
  \\
  UC Berkeley
}

\begin{document}
\maketitle

\vspace{-2em}
\begin{figure*}[h]
    \includegraphics[width=\linewidth]{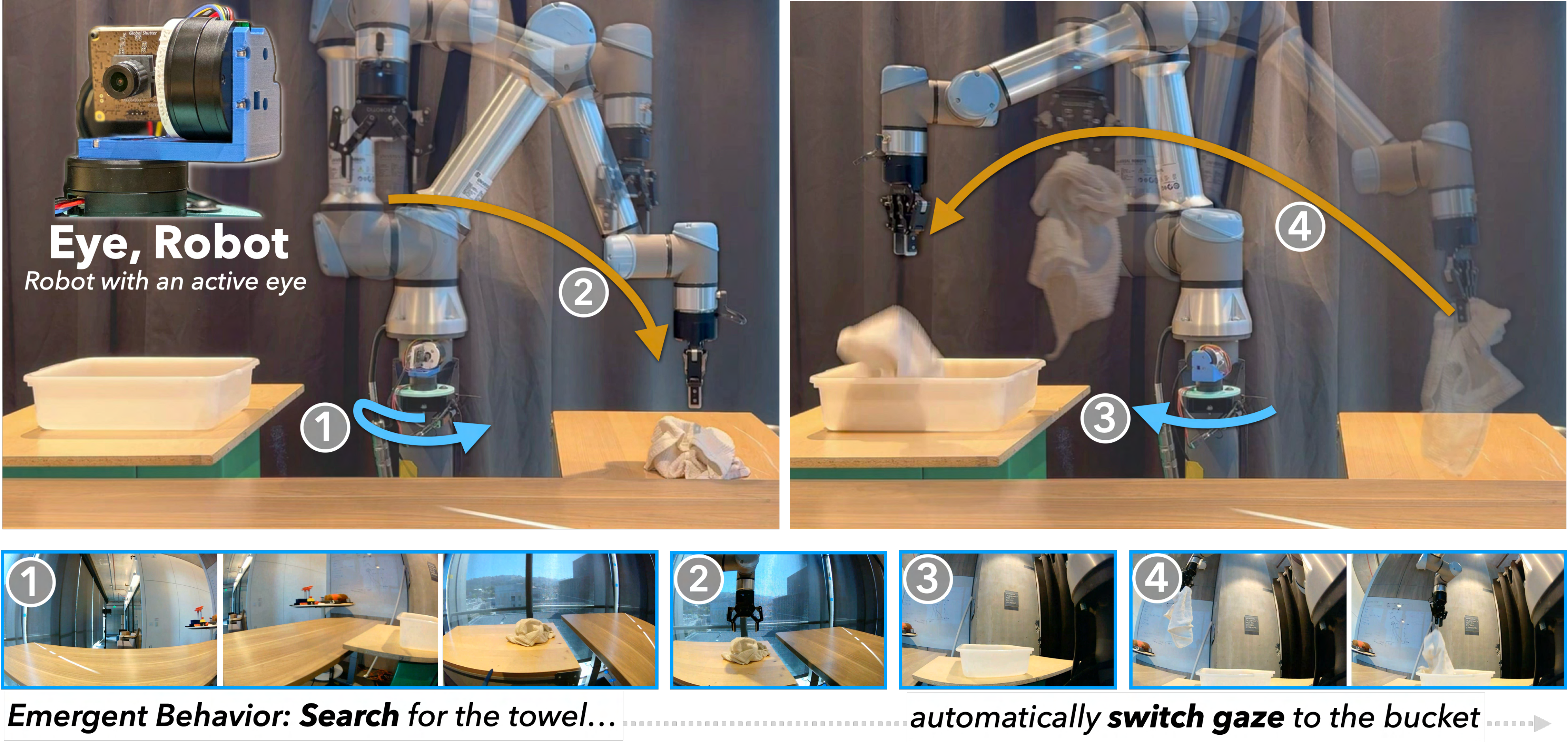}
    \caption{\textbf{EyeRobot.} We present a robotic system with an active eye, where the behavior of looking emerges from the need to act. A foveated mechanical eye, inspired by biological vision, is trained via reinforcement learning in a real-to-sim BC-RL loop. Shown here is a long-horizon pick-and-place task involving a towel and a bucket—neither visible in the initial view. The robot looks for the towel, grasps it, then switches gaze to the bucket to complete the task. This behavior emerges purely from being rewarded for looking in directions which facilitate action, without any gaze demonstration. %
    }
    \vspace{-1em}
    \label{fig:splash}
\end{figure*}
\begin{flushright}
\begin{minipage}{0.6\textwidth}
\itshape

\hspace{-12em}``We perceive in order to act and we act in order to perceive.'' \hspace{1em} --- JJ Gibson
\end{minipage}

\end{flushright}

\begin{abstract} 
Humans do not passively observe the visual world---we actively look in order to act.
Motivated by this principle, we introduce EyeRobot, a robotic system with gaze behavior that emerges from the need to complete real-world tasks.
We develop a mechanical eyeball that can freely rotate to observe its surroundings and train a gaze policy to control it using reinforcement learning.
We accomplish this by first collecting teleoperated demonstrations paired with a fixed $360^\circ$ camera. This data is imported into a simulation environment that supports rendering arbitrary eyeball viewpoints, allowing episode rollouts of eye gaze on top of robot demonstrations. We then introduce a BC-RL loop to train the hand and eye jointly: the hand (BC) agent is trained from rendered eye observations, and the eye (RL) agent is rewarded when the hand succeeds. In this way, hand-eye coordination emerges as the eye looks towards regions which allow the hand to complete the task.
EyeRobot uses a foveated vision transformer architecture, allowing high resolution with a small compute budget, which we find leads to the emergence of stable eye fixation as well as improved ability to track objects and ignore distractors.
We evaluate EyeRobot on five panoramic workspace manipulation tasks requiring manipulation in an arc surrounding the robot arm. Experiments suggest EyeRobot exhibits hand-eye coordination behaviors which effectively facilitate manipulation over large workspaces with a single camera.

\end{abstract}

\section{Introduction} 
\vspace{-0.5em}
\begin{figure}
    \centering
    \includegraphics[width=\linewidth]{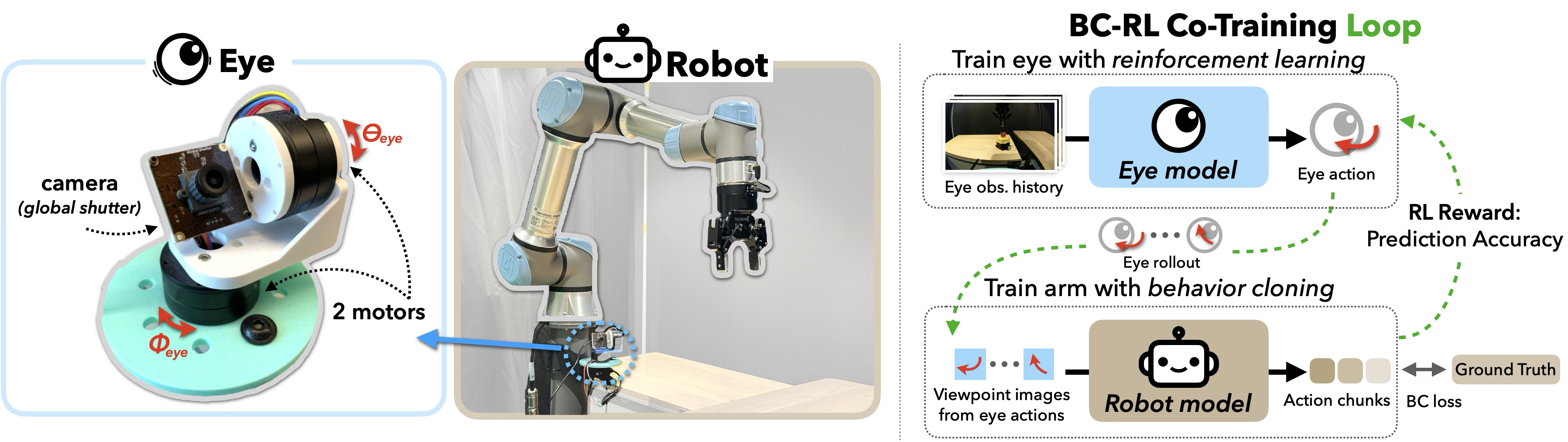}
    \caption{\textbf{EyeRobot framework.} Left: We develop a mechanical eyeball with two degrees of freedom, mounted on a high-speed gimbal and equipped with a fisheye lens and global shutter. Right: The eye policy is trained via a BC-RL loop which allows gaze to emerge to facilitate action. 
    }
    \vspace{-1em}
    \label{fig:hardware}
\end{figure}

Are you thirsty? Take a moment to reach for the nearest cup. As you do, your eyes move first—scanning the table, darting from region to region to locate the cup. Once it's in sight, your hand follows. This tight coupling between attention and eye movements is not incidental; it reflects a fundamental constraint of the human visual system. We are not built to perceive everything at once. As a result, we must look around---and in this sense, vision can be understood as a form of search. But what drives the search? The need to \textit{act}---to accomplish something in the world. Whether gathering information about task-relevant properties or guiding the execution of a precise action, we look because we have something to do. %

In this work, we present a robotic system where the same principle---looking to enable action---emerges naturally from a desired real world task without the need for gaze demonstrations, implemented on a mechanical eyeball that can freely rotate to observe its surroundings. The core challenge is how to train such a visual agent within the constraints of the physical world?
To address this, we introduce a behavior cloning (BC) - reinforcement learning (RL) loop enabled by a 360$^\circ$ camera real-to-sim environment.
This does not require any gaze demonstrations; instead, hand-eye coordination emerges solely from task supervision.

To specify what ``action'' we desire from the system, we build on recent advances in policy learning from behavior cloning~\cite{zhao2023learning}. %
We augment an existing teleoperation system to collect synchronized $360^\circ$ video and robot trajectory data, creating an EyeGym environment that enables replay of demonstrations with renderings from simulated eyeball viewpoints. %
Equipped with this environment, we propose a BC-RL algorithm for training an RL eye policy with a task-based reward, which samples rollouts from the current eye agent to use as observations to supervise a behavior cloning agent for the arm. The accuracy of these action predictions are cyclically used as rewards for the eye agent (Fig.~\ref{fig:hardware}). As these agents co-train, the eye thus learns to look around to optimize the performance of the behavior cloning agent---gaze emerges from the need to act. Figure~\ref{fig:splash} illustrates an example of search behavior which emerges in EyeRobot to accomplish a long-horizon pick-and-place task involving a towel and a bucket---neither of which is visible in the initial camera view. 

Inspired by nature, we design a Foveal Robot Transformer (FoRT) architecture which processes visual input in a foveated manner: peripheral vision provides broad, low-resolution coverage of the visual field, while foveal vision offers high-resolution input over a restricted area. Our experiments show this multi-resolution architecture results in enhanced fixation during task execution, distance invariance while tracking objects, and better ability to ignore distractors.
Our experiments evaluate EyeRobot on 5 large, panoramic-workspace manipulation tasks, involving objects on a $180^\circ$ area surrounding the arm. Across these tasks we observe a number of behaviors not explicitly trained for---gaze switching between targets depending on task stage, long-range search, and independently coordinated hand-eye movements. We find that EyeRobot enables promising performance on a variety of large-workspace pick-and-place and servoing tasks with \textit{only} a single egocentric active camera. 
We compare to externally and wrist-mounted cameras, and find that EyeRobot outperforms them in this large workspace either due to the limited resolution of the zoomed-out perspective or the inability of the wrist camera to search.

\section{Related Work}
\paragraph{Active Vision}

Active perception systems physically move sensors to not only see, but to \textit{look}~\cite{bajcsy1988active,bajcsy2018revisiting,goldberg1984active,aloimonos1988active}.
This arises naturally from physical constraints faced by embodied agents, which have only partial observations of the world at any point in time.
The key insight of active vision is that actuation lets agents shape the utility of these observations to better achieve their goals.
Existing systems primarily use active vision to maximize information gain for visual tasks: examples include tracking~\cite{papanikolopoulos1993visual,papanikolopoulos1991vision}, search~\cite{zhu2017target,ye2018active,kolve2017ai2}, observation completion or reconstruction~\cite{jayaraman2018learning}, and view selection for 3D reconstruction~\cite{zeng2020viewplanningsurvey,banta2000next,mendoza2020supervised,Smith2021Active3DVisionTouch,Jiang2023FisherRF} using voxel~\cite{isler2016infogainactive3d}, surfel~\cite{Monica18SurfelNBVPlanning}, or point cloud~\cite{Zeng2020PointCloudNBVPlanning} representations.
Other works couple active sensing setups with robot manipulators and evaluate systems in terms of downstream task performance: this includes improvements in semantic understanding~\cite{sripada2024ap}, computational efficiency~\cite{cheng2024open}, information seeking~\cite{dass2024learninglookseekinginformation}, and occlusion robustness~\cite{chuang2024active} for robot manipulation.
Like these systems, EyeRobot also studies active vision for robot manipulation.
Instead of relying on heuristics~\cite{sripada2024ap}, camera action demonstrations~\cite{cheng2024open,chuang2024active}, or ground-truth gaze or state however, EyeRobot proposes to learn optimal eye gaze policies directly in observation space with reinforcement learning; we learn where to look in order to act.

\vspace{-1em}
\paragraph{Behavior Cloning}
Behavior cloning (BC)~\cite{billard2008survey,hussein2017imitation,ravichandar2020recent,pomerleau1988alvinn} is the dominant paradigm for teaching robots manipulation skills.
BC is advantageous because it does not require hand-designed behaviors, costs, and rewards---instead, BC policies are simply trained to imitate human demonstrations.
Prior work has shown how this can enable new capabilities in mobile~\cite{du2022bayesian,fu2024mobile}, bimanual~\cite{zhao2023learning,lin2024learning,grannen2023stabilize,chi2024universal}, and language-conditioned~\cite{brohan2022rt,brohan2023rt,black2410pi0,team2025gemini} robot manipulation.
Policies of this form can also be implemented using a diverse range of architectures, including energy-based~\cite{florence2022implicit}, diffusion~\cite{black2410pi0}, and autoregressive~\cite{pertsch2025fast,team2025gemini}.
In EyeRobot, we adopt a similar imitation-based system for manipulation in large workspaces.
In contrast to prior systems that solely focus on learning from demonstration, however, EyeRobot's BC-RL training shows how behavior cloning objectives for robot actions can (i) be optimized jointly with an eye gaze reinforcement learning objective and (ii) used as a reward itself for reinforcement learning.

\vspace{-1em}
\paragraph{Biology-Inspired Machine Perception}
Many computer vision systems draw inspiration from biology to improve efficiency, adaptability, and performance.
For example, simple stereo cameras~\cite{marr1991computational,kanade1996stereo,scharstein2003high} mimic the binocular vision of animals to support depth perception, while event cameras~\cite{rebecq2019high,gallego2020event,kim2016real} mimic retinal spikes to enable low-latency perception.
Foveated systems emulate the resource-rational nonuniformity of the retina, using either specialized hardware~\cite{bandera1989foveal,minut2000face,killick2023foveation} or learned neural mechanisms~\cite{jonnalagadda2021foveater,cheung2016emergence}.
Beyond sensing, gaze control has been modeled after oculomotor behaviors like smooth pursuit and saccades~\cite{rivlin2000control}, and implemented in hybrid pipelines that combine peripheral detection with foveal tracking~\cite{gould2007peripheral}.
Given plentiful simulation, prior works have found foveation emerges when evolving a visual sensor to complete classification or survival tasks~\cite{tiwary2025eyecomputationallyrecreatingvision,cheung2017fovea}.
EyeRobot builds on the same principles as these prior works, drawing on the benefits of foveation and gaze control.
However, its implementation differs significantly: rather than rely on specialized sensors or hardcoded behaviors, we mount a standard RGB camera on a high-speed gimbal, apply foveation using a multi-resolution transformer, and learn a gaze policy via reinforcement learning with a real-to-sim gaze rendering pipeline.

\section{Approach}
\begin{figure*}[t]
\centering
\includegraphics[width=\textwidth]{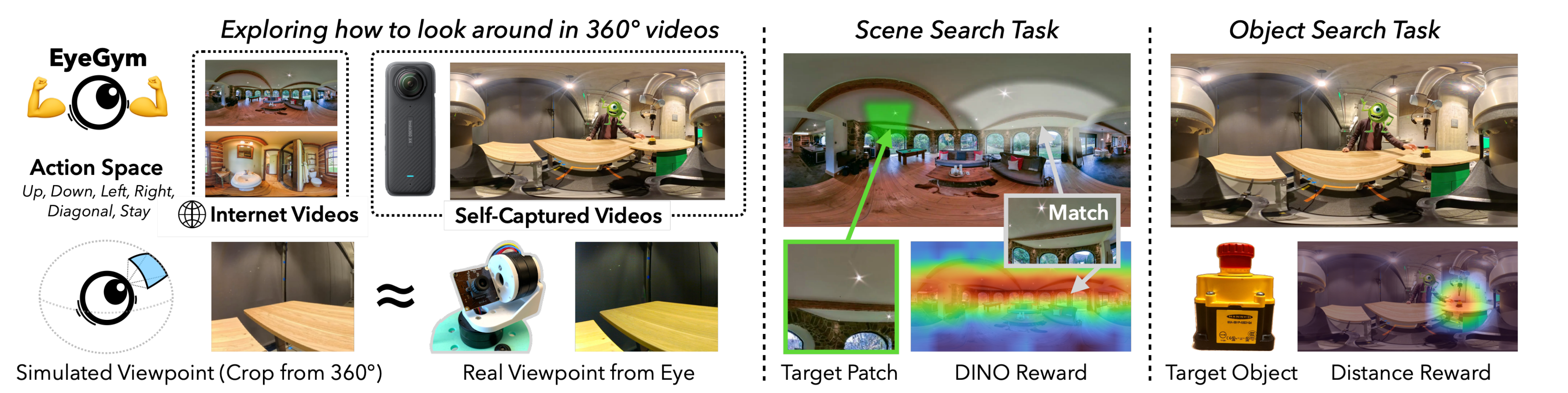}
\caption{\textbf{Learning to Look with EyeGym.} EyeGym enables training policies on 360° internet images or robot data to perform semantic visual search tasks. }
\vspace{-1em}
\label{fig:eyegym}
\end{figure*}

EyeRobot trains gaze policies for manipulation using reinforcement learning.
We conduct experiments using a UR5e robot arm with a gimbal-mounted camera mounted rigidly to the base of the robot.
To train gaze policies for hand-eye coordination, we present (i) EyeGym, an environment for eye gaze simulation, and (ii) reinforcement learning methodology---visual search rewards and a joint BC-RL loop---for learning gaze policies.

\subsection{Scalable Experience Collection with EyeGym}
\label{sec:real2sim}

Reinforcement learning benefits from scalable experience collection. %
To facilitate this for eye gaze policies, we introduce EyeGym: an RL environment that simulates eye gaze by sampling from $360^\circ$ image and video data.
Unlike prior approaches that rely on synthetic environments and physics simulators to render simulated views~\cite{coumans2020pybullet,makoviychuk2021isaac,todorov2012mujoco,zakka2025mujoco,savva2019habitat,szot2021habitat,puig2023habitat}, EyeGym renders directly by sampling from real equirectangular videos and images. %
This has several key advantages: (i) it reduces the sim2real gap by exposing policies to authentic textures, lighting, and noise; (ii) it enables training on native $360^\circ$ datasets~\cite{chou2020360,wallingford2024image,hu2017deep}, making EyeGym practical and scalable for learning active visual perception policies that transfer to real-world camera systems (Fig.~\ref{fig:eyegym}); and (iii) it incurs less overhead than traditional rendering.
We will release EyeGym code to support further research.

We use EyeGym for robot manipulation by first replacing our robot's physical eyeball with an off-the-shelf Insta360 X4 $360^\circ$ camera.
We can then use this camera to capture robot demonstrations teleoperated using a GELLO system~\cite{wu2024gello}, where the 5.7K, 30FPS equirectangular video sequences are recorded and synchronized with robot trajectories. %
Paired data can then be imported into EyeGym for simulating eye gaze on top of demonstrated robot motion. Advantages of this setup include minimal additional hardware, minimal additional data bandwidth, and compatibility with existing teleoperation systems~\cite{zhao2023learning,wu2024gello}.

\subsection{Learning Gaze Policies with BC-RL}

The goal of EyeRobot is to learn gaze policies that improve downstream task performance.
This is done without expert gaze demonstrations.
We instead use the EyeGym environment to learn gaze policies from flexible reward signals: either task-driven BC-RL or pure visual search.

\textbf{The BC-RL Loop}
The utility of gaze policies can ultimately be measured by downstream task performance.
Therefore, we propose to optimize such policies with robot task metrics, rather than hand-designed rewards like semantic visual search or user-specified gaze demonstrations.
We propose a BC-RL loop to achieve this.
Given a BC policy controlling a robot arm and an RL policy controlling gaze, the key idea of BC-RL is that observations flow from the RL policy to the BC policy, while task success metrics flow from the BC policy to the RL policy.
At every BC-RL optimization step, the eye policy attempts to optimize the BC agent's \textit{current performance} at matching demonstrated actions, while the BC agent learns how to perform the task best given the current gaze behavior.

We operationalize BC-RL in EyeGym using the synchronized demonstration and 360$^\circ$ video pairs detailed in Section~\ref{sec:real2sim}.
One episode rollout consists of a video trajectory sampled from a demonstration; beginning with the eye gaze randomly initialized $\pm90^\circ$ and $\pm15^\circ$ from the neutral azimuth and elevation positions.
At each step, the eye receives a reward comparing the predicted action chunk to the ground-truth action chunk. We use a reward based on the action chunk's forward-kinematics, which is better normalized than joint-space error. Specifically, we compute the end effector position over predicted and ground-truth trajectories, and assign reward to be the negative Fréchet distance between these splines (penalizing deviation). At the beginning of each demonstration, we pause time for 30 frames without BC supervision, to allow the eye to visually search before advancing time in the video. The BC agent is trained with a standard L1 loss between predicted and ground truth chunks.

\textbf{Active Visual Pretraining (AVP)}
Randomly initializing networks during BC-RL results in an ill-conditioned BC objective, as often times the target object is invisible given random eye movements.
We thus use a visual object search reward as a pretraining task to initialize network weights and ease BC-RL convergence.
We pretrain policies in EyeGym using randomly sampled static 360$^\circ$ video frames, allowing the eye to look around and rewarding it when the object is centered in view. During pretraining, we condition on CLIP embeddings of search targets to give networks the ability to learn about multiple objects; this input is replaced with zero tokens after pretraining. During the BC-RL phase, we initialize network weights of both the hand and eye with pretrained weights from AVP.

\begin{figure}
    \centering
    \includegraphics[width=\linewidth]{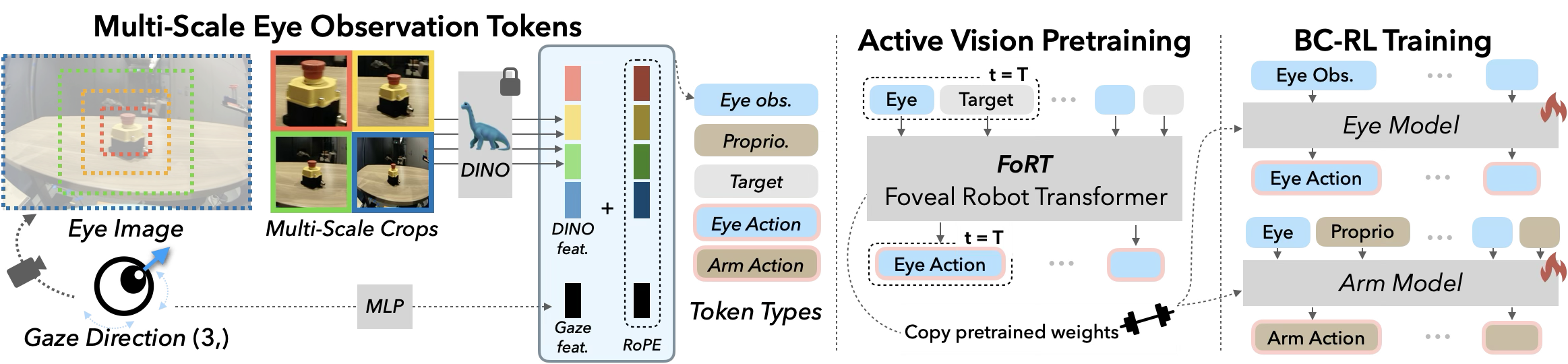}
    \caption{\textbf{Foveal Robot Transformer (FoRT)}. Eye observations are processed in a \textit{foveal} manner, where each image is processed at multiple scales and concatenated together with the gaze direction.}
    \label{fig:architecture}
    \vspace{-1em}
\end{figure}

\begin{wrapfigure}{r}{0.4\textwidth}
\vspace{-4em}
\includegraphics[width=\linewidth]{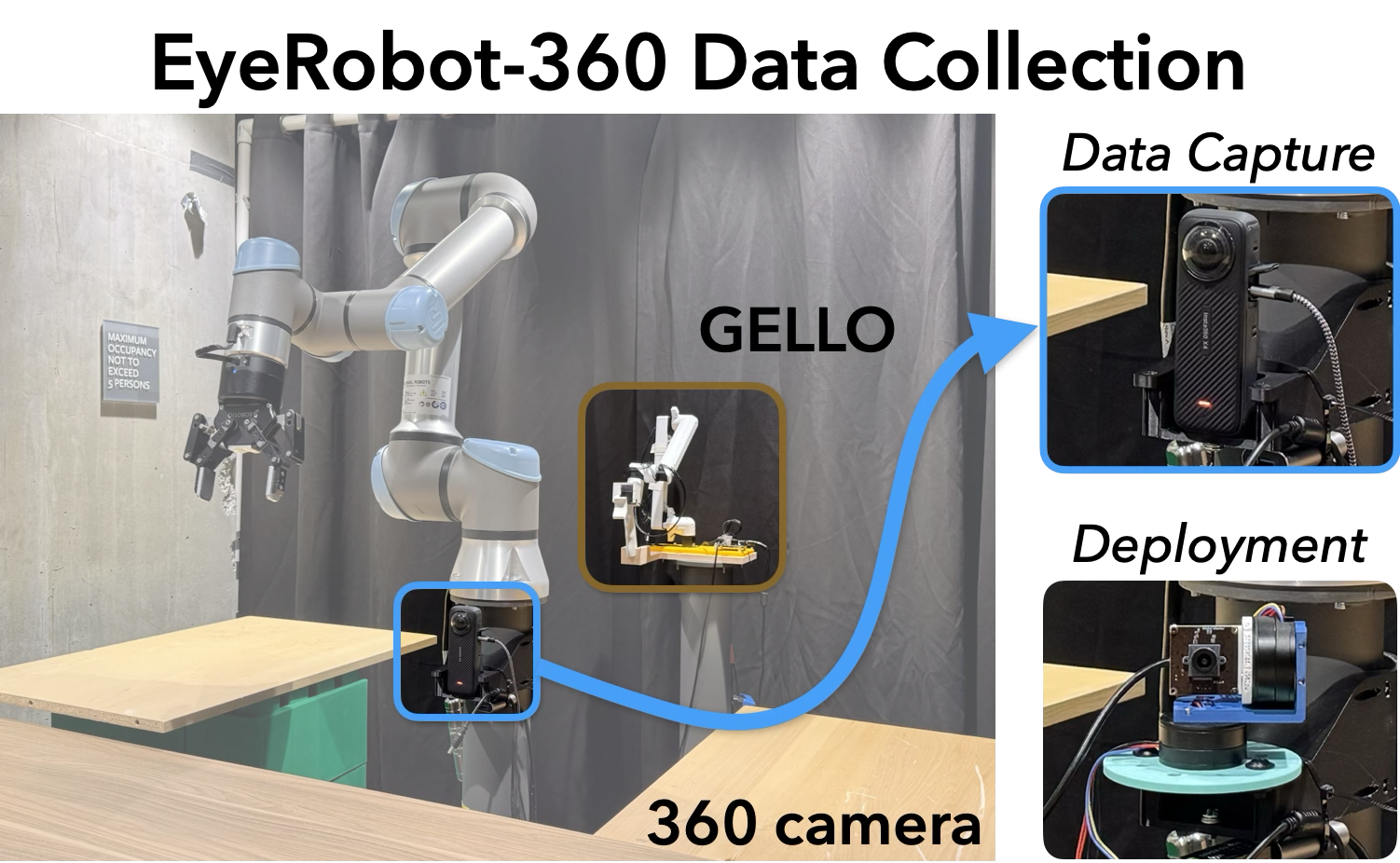}
\caption{\textbf{Teleop Data Setup with EyeRobot.} We collect actions with GELLO~\cite{wu2024gello}, and the EyeGym environment with a time-synced 360 camera.}
\vspace{-1em}
\label{fig:wrapfig}
\end{wrapfigure}

\subsection{Foveal Robot Transformer (FoRT)}
\label{sec:fort}
EyeRobot uses transformer architectures for its eye and robot policies (Fig.~\ref{fig:architecture}), which convert all observations to tokens and predicts all outputs as tokens.
The eye and robot policies share projection matrices for shared inputs, but otherwise use separate transformer weights.

\textbf{Observation Feature Extraction} Though sophisticated mechanisms for multi-resolution foveation have been proposed in prior vision work \cite{jonnalagadda2021foveater, deza2020emergent}, we wish to leverage pretrained vision backbones and thus opt for a simpler architecture involving multi-cropping. We process input images into an image pyramid of crops with N scales centered at the center pixel, rescaling all images to the same 224 resolution. We embed all patches independently with a frozen DINOv2-ViT/S~\cite{oquab2023dinov2} encoder, and flatten them token-wise as inputs to the transformer. Each policy additionally takes as input the eye gaze direction as a 3D vector, the current joint proprioception, and an optional ``target" token for visual search. Each are projected to the input token dimension with small MLPs. We apply 10\% dropout on the proprioceptive tokens to avoid overfitting.  

\textbf{Outputs}
Action outputs are decoded from the transformer with lightweight projection heads. Eye actions are parameterized as a categorical distribution over 8 azimuth-elevation directions and $\vec{0}$. %
We use a learnable input token for each output type, which is shared across timesteps in the output action chunk.

\section{Experiments}
Our experiments aim to (1) evaluate the ability of using RL to visually search in scenes and find objects, (2) compare EyeRobot's performance on manipulation tasks to conventional camera placements, and (3) investigate emergent properties of the hand-eye coordination learned during BC-RL.

\subsection{Using EyeGym for Visual Search}
Here we perform an RL-only experiment to evaluate the effectiveness of the EyeGym at training an eyeball to search the environment for a desired object.
\begin{wrapfigure}{r}{0.4\linewidth}
    \vspace{-1.5em}
    \centering \includegraphics[width=\linewidth]{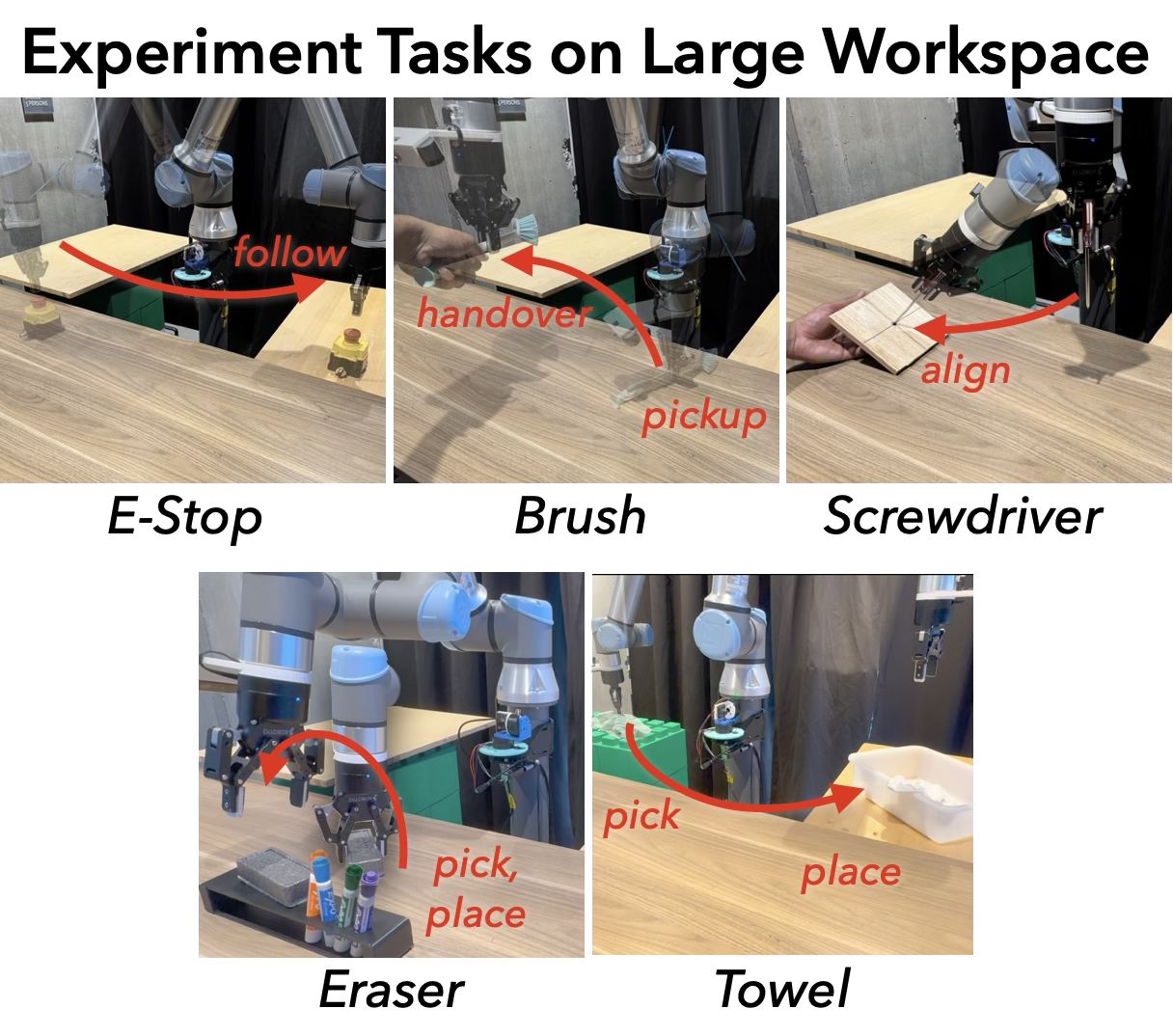}
    \caption{\textbf{Tasks.} We evaluate EyeRobot on 5 large-workspace tasks.}
    \vspace{-2em}
    \label{fig:tasks}
\end{wrapfigure}
\textbf{Visual Search Rewards}
Search is a critical step of vision for robot manipulation, especially for large workspaces.
As an initial gaze policy study, we evaluate two search tasks with explicit visual rewards.
\textit{(a) Scene Search} policies attempt to locate image patches that are visually similar to a given target patch.
For this, we use DINO feature similarity between current and target views as a reward signal. See the Appendix for results and discussion.
\textit{(b) Object Search} policies are more fine-grained, and aim to locate specific objects within scenes.
For Object Search, we primarily investigate a ``truncated distance reward''. This reward is zero if the target object is outside the camera's field of view (FOV) and increases linearly to 1 as the agent perfectly centers the target in its view.

\textbf{Object Search}
We define a task where the goal is to locate and track target object(s). We investigate this setting using self-captured videos from robot demonstrations, train with the “Truncated Distance” reward, and deploy the learned policy to the physical eyeball. We train the robot to locate the towel and assess its ability to find it. The towel is deliberately placed outside the initial field of view, requiring the eyeball to actively explore the scene. We achieve a success rate of 87\% with an average search time of 1.8 seconds, with failures occurring when the towel lies near the far edges of the workspace. To investigate qualitative tracking speed capabilities, we also train on a simple stuffed bear, and find the eye can track even rapid movements like tossing, see project site for videos.

\begin{figure}
\centering
    \includegraphics[width=0.9\linewidth]{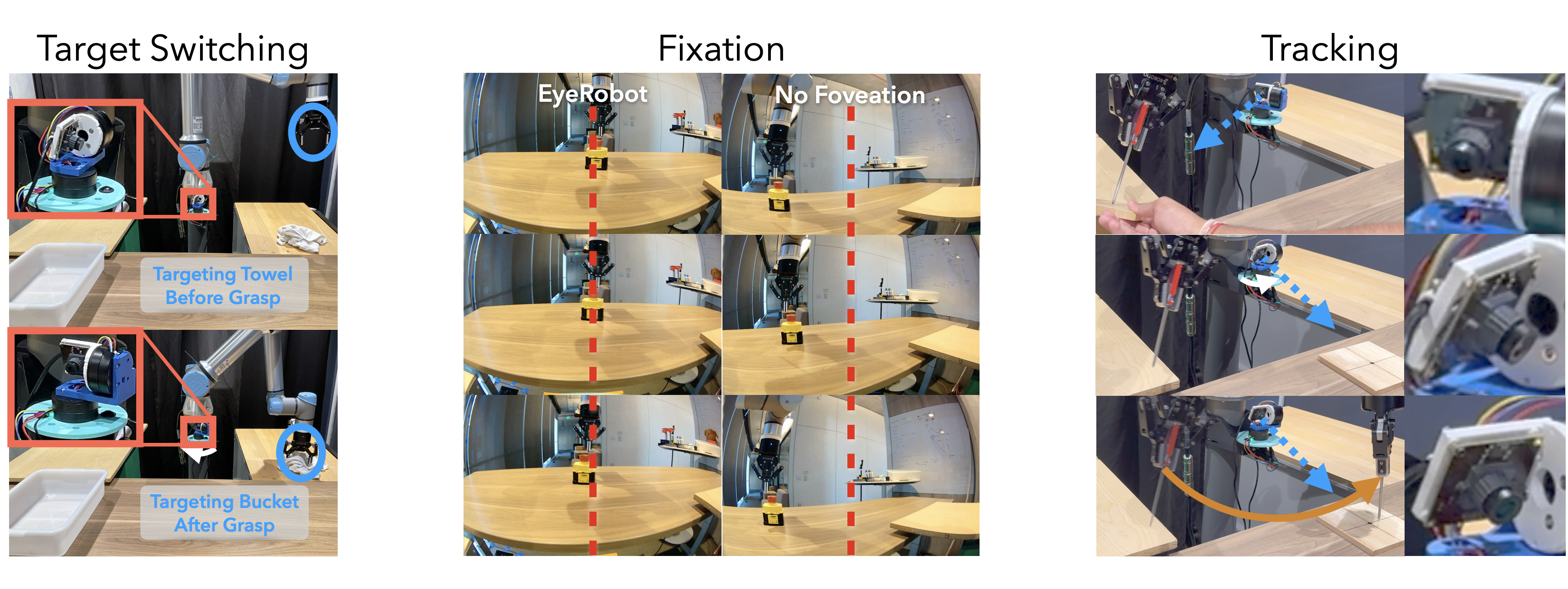}
    \caption{\textbf{Emergent Eye Behavior.} (Left) The eye learns to shift its gaze from towel to bucket during grasping depending on the state of the arm. (Middle) Foveation causes object-centering fixation to arise. (Right) Gaze follows target objects independently of the hand. }
    \label{fig:interesting}
    \vspace{-1em}
\end{figure}

\subsection{Hand-Eye Coordination}
We evaluate EyeRobot on 5 tasks (Fig.~\ref{fig:tasks}) which involve manipulation over a $210^\circ$ workspace, to probe the limits of hand-eye coordination over a large region. All experiments are performed on a UR5e with a Robotiq gripper. Data was collected using GELLO~\cite{wu2024gello}, totaling 100-500 demos for each task. This workspace represents a significant challenge: some tasks require tolerance on the order of $cm$, while the workspace is on the order of $1000cm^2$. For all trials, the robot is programmatically reset to the same pose (shared across ablations and baselines). We evaluate on 5 large-workspace manipulation tasks: eraser-on-shelf, e-stop reaching, brush handoff, screwdriver servoing, and towel-in-bucket. See Appendix for details on these task definitions.

\textbf{Results} Results are reported in Tables ~\ref{tab:baselines}, ~\ref{tab:estop_ablations}, ~\ref{tab:towel_ablations} and results are best viewed in the included execution videos to better understand qualitative performance. EyeRobot can consistently perform manipulation tasks over a large workspace, For the \textbf{Towel} task,  Most of EyeRobot's failures are in narrowly missing grasps on the towel, and we notice it tends to struggle more in switching gaze from bucket to towel, as evidenced by worse performance in trials where only the bucket is visible. Sometimes it grasps only a corner of the towel, leading to a difficult bucket drop where half the towel dangles outside the bucket. In our \textbf{E-Stop} trials, EyeRobot never loses track of the object, and its main error is in z-distance towards the eyeball, owing to the difficulty of resolving depth with a monocular viewpoint. In the \textbf{Eraser} task, EyeRobot primarily fails by narrowly missing grasps on the eraser. In the place only trials where the eraser begins in the same location each time, EyeRobot can robustly follow the perturbed location of the shelf post-grasp. 
In \textbf{Screwdriver}, EyeRobot achieves a mean error of 4.0cm when the target is flat, and 5.2cm when the target is tilted $45^\circ$, compared to a total test area spanning 115cm left to right. The \textbf{Brush} task successfully grasps the brush at the correct orientation 15/20 trials, and successfully completed the human handover 14/15 times.

\begin{wrapfigure}{r}{0.4\linewidth}
    \vspace{-1.5em}
    \centering \includegraphics[width=\linewidth]{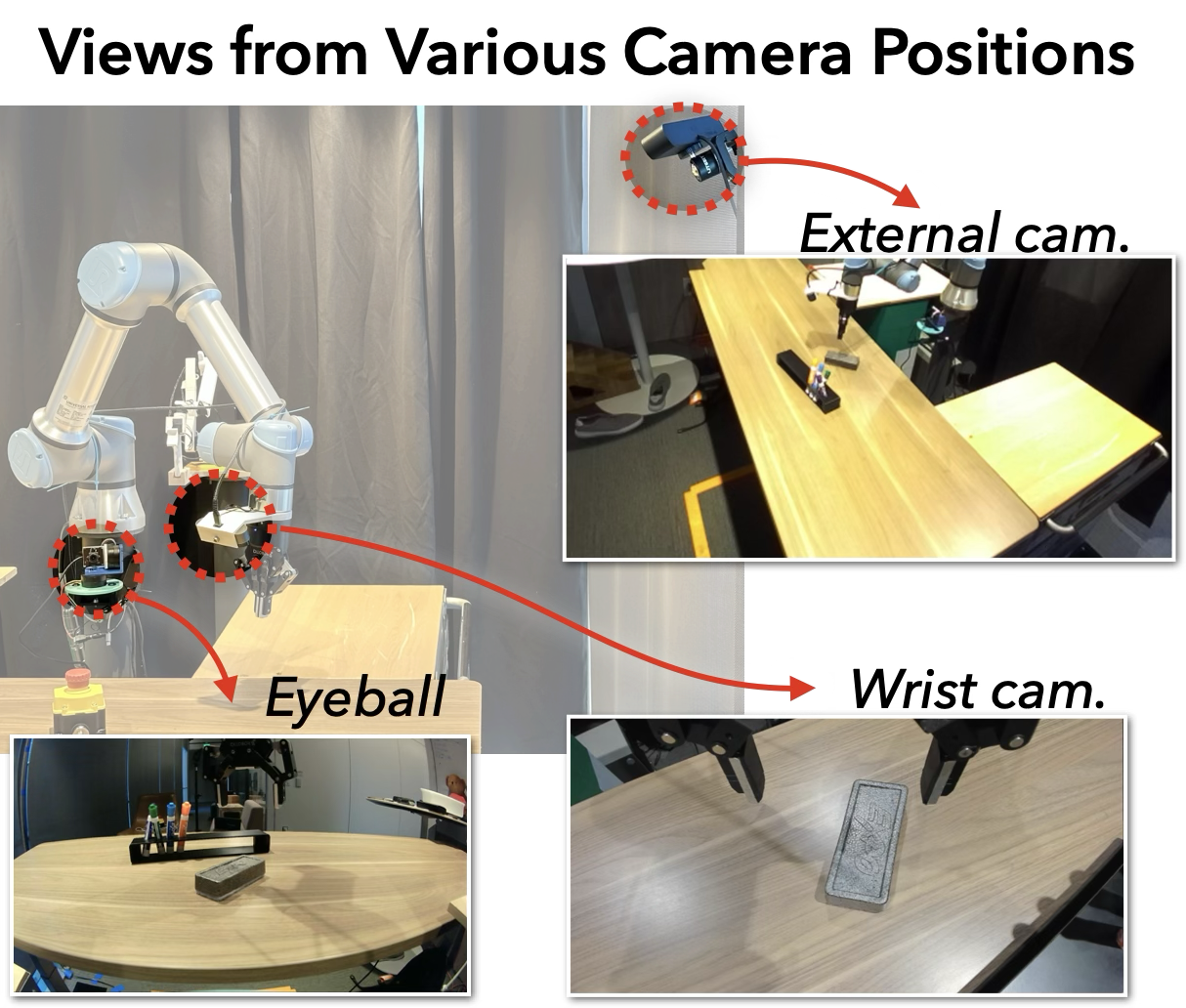}
    \caption{\textbf{Effect of Camera Placement.} Placing the camera on a gimbal allows it to observe across the whole workspace at a higher resolution.}
    \label{fig:views-from-cams}
\end{wrapfigure}
\subsubsection{Eye Behavior}

We observe three key behaviors that the eye learns while being rewarded for enabling action: switching, search, and independent tracking. In multi-step tasks, the eye learns to automatically switch its gaze towards the next relevant object, depending on the state of the robot (Fig~\ref{fig:splash}). For long-horizon tasks this requires that the eye search for the subsequent target objects when it is not in view, or even just make smaller fixation adjustments for objects that are only partially visible (Fig~\ref{fig:interesting}). Its search strategy tends to oscillate back and forth in a sweeping motion. When the target location is moved mid placement (e.g., `perturb' condition in Table 2) the eye tracks the new target location, independently of the robot arm. Finally, in more dynamic tasks, we note that the eye learns to attend to predictive cues in the environment (e.g., humans placing a target object). Taken together, these qualitative results suggest the potential of an active vision agent trained with RL to naturally complement a behavior cloning agent in achieving tasks.

\subsubsection{Wrist, Exo Comparisons}

\begin{table*}[t]
\centering
\scriptsize
\captionsetup{font=footnotesize}
\caption{\textbf{Camera Comparisons}. We report success rate (Eraser \& E-Stop) and distance to target (E-Stop).}
\label{tab:baselines}
\setlength{\tabcolsep}{3pt}
\begin{tabular}{lcccc | @{\hspace{10pt}}lcccc}
\toprule
Task & EyeRobot & Exo & Wrist & Wrist+Exo & Task & EyeRobot & Exo & Wrist & Wrist+Exo \\
\midrule
Eraser (Pick \& Place) & 60\% & 0\% & \textbf{100\%} & 60\% & Eraser (Perturb Place) & \textbf{100\%} & - & 10\% & 40\% \\
E-Stop (Slow) & \textbf{100\%} & \textbf{100\%} & 80\% & \textbf{100\%} & 
E-Stop (Fast) & \textbf{100\%} & \textbf{100\%} & 60\% & \textbf{100\%} \\
E-Stop (Slow) & \textbf{4.0cm} & 7.8cm & 5.3cm & 7.1cm & 
E-Stop (Fast) & \textbf{4.7cm} & 9.9cm & \textbf{4.7cm} & 5.5cm \\
\bottomrule
\end{tabular}

\vspace{-1em}
\end{table*}

An important question is how EyeRobot compares to more conventional camera placements, namely fixed externally-mounted, and wrist-mounted cameras (Fig.~\ref{fig:views-from-cams}). We perform comparisons on the E-Stop and Eraser tasks, by training on \textit{the exact same data} with the same architecture. For exo-only and wrist-only baselines, we input images at 640 resolution, while for wrist+exo we use 360. Both comparisons have \textit{more} input image tokens than FoRT, and the same number of model parameters.  

\textbf{Results} are reported for the eraser and e-stop tasks in Table~\ref{tab:baselines}. When the wrist camera can view the eraser, the wrist camera achieves a very high success rate, although this performance greatly suffers in the perturbation trials where the wrist cannot search. In contrast, EyeRobot can adjust its viewpoint to maintain the shelf in view at all times. The exo camera baseline struggles to achieve precise enough grasps given its low resolution; while often touching the eraser, it never successfully grasps it.
The exo+wrist baseline also performs worse at servoing compared to wrist-only, likely due to the complexity of merging image tokens from multiple cameras--though it can occasionally succeed in locating the perturbed shelf.

\subsection{Ablations}

\begin{wraptable}{r}{0.48\textwidth}
\vspace{-2em}
\raggedleft
\scriptsize
\captionsetup{font=footnotesize}
\caption{\textbf{Foveation Ablation.} (Error {\scriptsize\(\downarrow\)} / Speed {\scriptsize\(\downarrow\)})}
\label{tab:estop_ablations}
\begin{tabular}{lcc}
\toprule
Task & EyeRobot & No Foveal \\
\midrule
E-Stop (Slow) & \textbf{4.0cm} / \textbf{4.2s} & 5.9cm / 5.3s \\
E-Stop (Fast) & \textbf{4.7cm} / \textbf{5.8s} & 6.9cm / 7.1s \\
\midrule
Average & \textbf{4.4cm} / \textbf{5.0s} & 6.4cm / 6.2s \\
\bottomrule
\end{tabular}
\vspace{1em}
\centering
\scriptsize
\captionsetup{font=footnotesize}
\caption{\textbf{AVP Ablation.} (Success Rate)}
\label{tab:towel_ablations}
\begin{tabular}{lcc}
\toprule
Visible At Start & EyeRobot & No AVP \\
\midrule
Both Visible & \textbf{80\%} & 73\% \\
Towel Visible & \textbf{95\%} & 40\% \\
Bucket Visible & 50\% & \textbf{70\%} \\
Neither Visible & \textbf{60\%} & \textbf{60\%} \\
\midrule
Average & 
\textbf{72.2\%}{\scriptsize(±13.1\%)}
&
\textbf{62.1\%}{\scriptsize(±14.2\%)}
\\
\bottomrule
\end{tabular}
\end{wraptable}
\textbf{Foveation:} To understand the contribution of foveated inputs on EyeRobot's behavior, we train and evaluate a BC-RL gaze model that operates over a uniform image resolution. To control for the number of image tokens we use inputs of 1x448x448 instead of 4x224x224. 
We find the foveated model consistently maintains the target object near the center of its field of view, whereas the non-foveated model only loosely keeps it within view 
(Fig.~\ref{fig:interesting}). Quantitatively, this uniform image resolution leads to degraded performance on all metrics evaluated for E-Stop, while settling significantly slower due to its unstable viewpoint (Table~\ref{tab:estop_ablations}). 

The foveated model is also significantly more robust to distractor objects; when a human places a distractor (yellow cup) in the scene, the uniform resolution model tends to split attention between the two and pays more attention to the human hand, and by slowly moving the cup one can force the robot to completely lose the E-Stop. On the other hand, the foveated model is able to ignore this distractor and remains fixated on the E-Stop (see videos).
These data suggest potential performance benefits in adopting a foveal architecture for manipulation owing to the physical token distribution making it easier to attend to the right region. 

Foveation also results in significantly improved robustness to objects' distance during visual tracking. This phenomenon is best seen in experiment videos; while slowly moving the stuffed bear away from the eyeball, the uniform resolution model quickly loses track after about 2 meters as the bear vanishes from input tokens, whereas the foveated model can maintain tracking much farther (around $5\times$) because its pyramidal representation maintains tokens visually recognizable as the bear.

\textbf{Active Visual Pretraining}: We ablate AVP on the towel task, and find that long-range visual search can emerge \textit{purely} from task driven BC-RL. We note, however, that grasping performance suffers compared to the model with AVP, which we attribute partially to poorly initialized network weights and partially to poor training data distribution early in BC-RL training, as the eye fails to fixate correctly at initialization. Convergence speed of BC-RL is also aided by AVP as it initializes gaze looking towards task relevant objects.

\section{Conclusion}
EyeRobot presents a method for training a mechanical eyeball to look around to achieve physical robot manipulation via a real-to-sim-to-real pipeline utilizing $360^\circ$ videos. This simulation environment allows the training of active visual policies with RL on top of teleop demonstrations, with which we train a visual agent to look around to maximize task-based BC performance. We find this agent learns to look to facilitate action, resulting in emergent eye behaviors such as search and fixation, and that the eye enables manipulation across a large workspace. In addition, we find that physically allocating tokens towards target objects via foveation results in improved robustness to distractors as well as increased robustness to distance while maintaining a cheap compute budget.

\section{Limitations}
The primary limitation of EyeRobot is the inability of 360$^\circ$ video to simulate motion parallax, i.e, how a neck can provide the ability to move around occlusions for a better perspective. %
Another drawback of EyeRobot is its eagerness to learn strategies which match simulation very well, but fail in real due to narrow data distributions. Concretely, one behavior we observe in the towel task is ``blind grasping", where the robot will sometimes look left, and upon observing no towel, grasps an average location to the right. This arises from the demonstration data distribution, owing to the fact there are fewer demos at the edges of the workspace. Finally, training BC-RL converges significantly slower than vanilla BC.
This is because of the co-training of two models which are mutually used in the others' train loop.
While our current system is limited to a stationary workstation, an exciting future direction could be to mount the eyeball on a mobile robot, further increasing the need for active vision.  Likewise, adding a second eye for stereo vision or incorporating monocular depth estimators could help overcome current limitations in depth perception and support more fine-grained manipulation.

\section{Acknowledgements}
This research was performed at the AUTOLAB at UC Berkeley in affiliation with the Berkeley AI Research (BAIR) Lab, which is supported in part by donations from Bosch, Google, Autodesk, and Siemens. Justin Kerr, Kush Hari, and Chung Min Kim are supported by the National Science Foundation Graduate Research
Fellowship Program under Grant No. DGE 2146752. Any opinions, findings, and conclusions or
recommendations expressed in this material are those of the author(s) and do not necessarily reflect
the views of the National Science Foundation. The authors were supported in part by equipment
grants from NVIDIA. We deeply appreciate Jon Kuroda, Ion Stoica, and
Joey Gonzalez for their generous support with computing hardware during some downtime on compute.

\clearpage

\bibliography{example}  %
\clearpage

\appendix

\author{
  Anonymous submission \\
}

\section{Eye Hardware}
\label{sec:eye}
The physical eye uses a 90fps, 1900x1200 global shutter RGB camera with a $110^\circ$ FOV fisheye lens, which allows for wide periphery view, with smooth and unwarped images during rapid movements. The eye is mounted on a gimbal consisting of two direct-drive brushless DC motors independently controlling pan and tilt, which are capable of a stopped-to-stopped saccade speed of $20ms$ at a range of $60^\circ$. In practice, we limit and smooth the speed of motion during inference to minimize motion blur. The rapid motion capability of the eyeball aids sim2real transfer by reducing the gap between commanded and executed motion. CAD and hardware instructions will be made available.

\section{Task Descriptions}

\textbf{E-Stop Reaching:}
The robot must continuously servo its gripper to hover above the button at all times, tracking large perturbations up to $180^\circ$. We measure the settling time after each perturbation and metric error from tooltip to button. We perform one ``slow" trial where the E-Stop is moved to 10 locations sequentially spaced $20^\circ$ apart, and a ``fast" trial where these same locations are tested in an adversarial pattern up to $90^\circ$ apart.

\textbf{Screwdriver Servoing:} While grasping a 7cm screwdriver, the robot must servo to align its tip with the center of a piece of wood marked by a black dot, which can be oriented between flat on the table to $45^\circ$ inclined. This task requires orientation control and servoing of a very thin (3mm) tool shaft. Similar to the E-stop task, we measure final screwdriver tip distance for this task.

\textbf{Eraser on Shelf:} The robot grasps and transports an EXPO eraser onto a shelf, whose position may be perturbed during execution. The orientation of both pick and place locations can be randomized $\pm45^\circ$. We consider an experiment a success if the resting state of the eraser is fully supported by the shelf. We test on a static case, where both objects are static but their location varies $180^\circ$ radially around the robot, and a case where objects begin centered but we perturb the shelf location post-grasp by 40cm left or right. This evaluates the policies' robustness to searching for the target place location.

\textbf{Towel in Bucket:} The robot must transport a towel from the tabletop to a bucket, \textit{both of whose locations may vary} up to $180^\circ$. This task is the most difficult for visual search, as some cases (Fig.~1) begin with \textit{neither} target object in view. We split evaluation into 4 tiers with 10 trials each, corresponding to which objects are initially in view. We pre-define locations for each tier covering the entire workspace, and initialize the eye facing forward on cases involving search. A trial succeeds if the towel is fully inside the bucket when the bucket is lifted by the experimenter, with half credit assigned if the towel is partially inside the bucket when lifted.

\textbf{Brush Handoff:} The robot grasps a brush and servos it towards a human, releasing its grasp when the human grasps the handle. The brush orientation may vary $\pm90^\circ$, and its position is tested along a 75cm line in the center of the table. A handoff is considered successful if the robot picks up the brush, moves it within arms reach of the human, and releases its grasp once the human grasps the handle (but not before).

\section{Demonstration Data Details}
All data, code, and data collection hardware will be made public.
\subsubsection{Data Collection Methodology}
We collect teleop demonstrations for servoing tasks (E-Stop, Screwdriver) in one continuous take, while a human moves the target periodically around. For the towel task, data is collected with uniformly randomized positions of the towel and bucket, with care to ensure they are each placed everywhere around the workspace. For the Eraser task, the location of the eraser relative to the shelf is maintained roughly the same within a 20x20cm square, while the global positions of both vary around the whole workspace. In half of our demos, the positions of one or both of the eraser and shelf are perturbed by a human during execution to provide more rich servoing data to the robot. For the brush task, the brush is uniformly randomized on the front of the reachable workspace, and at each demo the human waits for a random amount of time before grabbing the brush for hand-off. This encourages the robot to learn when to release the brush rather than prematurely dropping it. 

\subsubsection{Dataset Statistics}
Table~\ref{tab:task_data} shows the size of our teleop datasets. These numbers account for the total amount of post-processed ``live" robot time, which the BC-RL agent sees during training.

\begin{table}
\centering
\begin{tabular}{lcr}
\toprule
Task & Number of Demos & Total Hours \\
\midrule
Eraser & 242 & 0.95 \\
E-Stop & -- & 0.50 \\
Screwdriver & -- & 0.71 \\
Towel & 599 & 1.69 \\
Brush & 265 & 0.83 \\
\bottomrule
\end{tabular}
\vspace{.2em}
\caption{Dataset size of teleop demonstrations. Continuous servoing tasks like E-Stop and Screwdriver were collected in a few, long demos, hence we only report total time.}
\label{tab:task_data}
\end{table}

\section{FoRT Implementation Details}
\subsubsection{Positional Embeddings}
We use 3D rotary positional embeddings (RoPE)~\cite{su2024roformer} to assign $(t,x,y)$ coordinates to each token. We assign each image token its corresponding image pixel coordinate, which allows attention to seamlessly transfer from one crop within the image pyramid to others. This allows the transformer to re-use features across scales, which provides a mechanism for correlating observations of the same object across distances. For example, an object close-up fills the periphery, while an object far away fills the fovea, allowing distance to be recognized via spatial attention while underlying token embeddings themselves remain similar. All other non-image tokens are assigned $(\frac{I_W}{2},\frac{I_H}{2})$ as their spatial coordinate to encourage preferential attention to the image center. Each $t$ coordinate is assigned to the corresponding timestep of the observation or output, for example a proprioceptive observation token at time $t$ would produce output joint tokens from $t$ to $t+A_\text{size}$
\subsubsection{Attention Masking and Memory}
EyeRobot masks significant portions of self-attention to increase the throughput of inference and minimize overfitting. First, we mask all image-image attention since DINOv2 outputs have already undergone self-attention. Second, we utilize sliding window attention to train on long rollouts efficiently (up to 100 time-steps), with a window size of 10 for the eye to provide history of motion, and 1 for the hand to effectively make it single-frame. 
We use Flex Attention~\cite{dong2024flexattentionprogrammingmodel} to efficiently compute these custom masks.

\section{BC-RL Implementation Details}

\subsubsection{Optimization and Hyperparameters}
We use PPO to optimize the eye agent during all training, with default parameters and an entropy regularization coefficient of $0.01$. We use separate optimizers for the hand and eye policies, both of which are optimized with AdamW with default parameters except a $\beta_2$ coefficient of 0.95. The learning rate is initialized to $1e-3$ for the BC agent and $5e-4$ for the RL agent, to incentivize the RL agent to move slower so the BC agent has more time to converge. Both learning rates following a cosine decay schedule to a $10\times$ lower LR, and are trained for 7M environment steps. We use an action chunk size of 30, corresponding to 1 second of real-world time. The BC agent outputs raw joint positions to execute (which are normalized to the range $[-1,1]$), which are parameterized either by absolute or relative joint commands. We find that for tasks with persistent large motions such as the E-Stop, Towel, or Screwdriver, absolute actions out-perform relative as the latter tends to get ``lost" in joint configuration space and has no data to recover from such states. In tasks with shorter horizon motion such as the Eraser and Brush, we find that relative actions can achieve more precise servoing performance.

\begin{wrapfigure}{r}{0.4\textwidth}
\vspace{-1em}
\includegraphics[width=\linewidth]{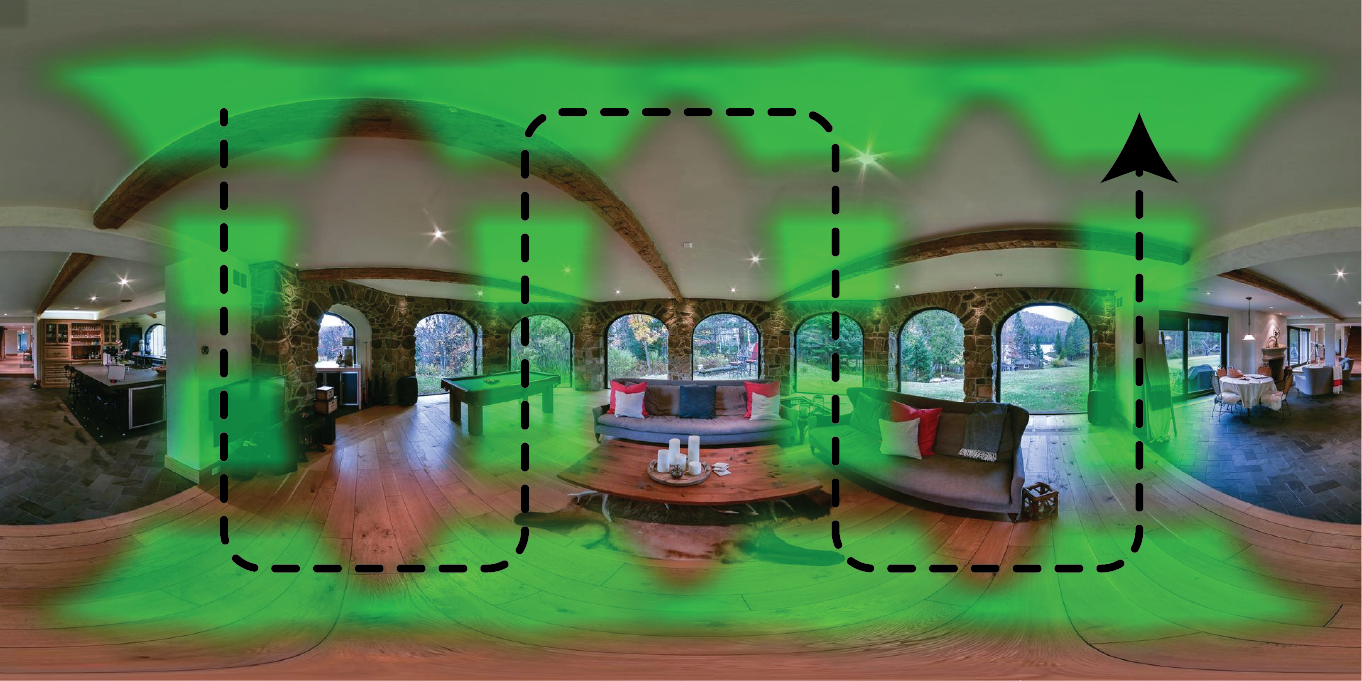}
\caption{\textbf{S-shaped pattern for scene search evaluation.} We move a target across crops in a sweeping pattern to probe our scene search behavior.}
\vspace{-1em}
\label{fig:movingtarget}
\end{wrapfigure}
\subsubsection{EyeGym Rollouts}
Each episode is rolled out for 130 steps, 30 of which are paused in the beginning to give the eye agent 1 second to explore and find relevant objects. The demonstration video is played at $2\times$ real-time speed to increase the diversity of batches seen by the BC agent. In effect, this means the agent sees 100 frames at 15fps, or 6.7 seconds of data each rollout. Each rollout is initialized at a random time-step from a randomly sampled demonstration, with an added buffer to ensure the rollout doesn't extend beyond the length of data available. We rollout 16 environments in parallel across GPUs, and use NVIDIA A6000s to train.

\begin{wraptable}{r}{0.45\textwidth}
\vspace{-2em}
\centering
\small
\caption{\textbf{Scene Search Results.}}
\label{tab:svs_metrics}
\begin{tabular}{lcc}
\toprule
Method & CLIP \(\uparrow\) & Exact Match \(\uparrow\) \\
\midrule
Random Walk          & 0.629 & 28.1\% \\
Distance Reward      & 0.704 & 64.8\% \\
DINO Reward          & 0.715 & 60.2\% \\
DINO+Distance        & 0.711 & 66.5\% \\
\bottomrule
\end{tabular}
\vspace{-1em}
\end{wraptable}

\subsection{Scene and Object Search}
\textbf{Scene Search}
We propose a scene-level semantic visual search task on 2,000 $360^\circ$ images from \cite{chou2020360}. For each training episode, we select a randomly located target crop with a field of view between 10° and 65° and condition FoRT on pooled extracted DINO tokens.
Evaluation is performed on 500 unseen images, each with 24 equally space target crops, and results are reported in Table~\ref{tab:svs_metrics}. We move the target in an S-shaped pattern, and for each new location, we allow the policy 20 steps to move before evaluating CLIP~\cite{radford2021learning} similarity and ``Exact Match'' rate, which marks how often it can put the target object within its field of view at the end of the rollout. We find the eye learns to look towards semantically similar viewpoints, but struggles to find exact matches due to the inherent repeating nature of many scenes.

We train policies with DINO feature similarity and truncated distance rewards and look for interesting emergent behavior. The results are best visualized in our video, where the policy learns to find semantically similar viewpoints. To test this behavior, we move a target through 360° images in an S-shaped pattern, as shown in Figure~\ref{fig:movingtarget}. The scene is divided into 24 crops arranged in a 4×6 grid (4 along elevation, 6 along azimuth), with each crop covering a 40° field of view. The target starts at the top left, and the policy has 20 steps to move around and look for similar-looking regions before the target moves to its next position, at which point it again has 20 steps. This setup allows us to observe visual search behavior. We evaluate this procedure across 500 images and report CLIP similarity and exact match rate, computed at the final step after the policy has used its 20 steps to move. Our numbers are in the main paper, where we compare our method variants against a random movement baseline. We chose the truncated distance reward for active visual pretraining since it leads to more search behavior, where the policy is only rewarded when it can find the target object.

For the object search experiment using our real camera to find the towel, we run 15 trials: 5 with the towel placed on the right, 5 in the center, and 5 on the left. In each case, we initialize the eyeball looking away from the towel to ensure a non-trivial starting point. The model successfully finds the towel in 87\% of the trials, as reported in the main paper, indicating it has learned a valid search behavior that can help initialize the policy for active vision pretraining in robotic manipulation tasks.

\subsection{Robot Inference}
When deploying on the robot, we predict eye and arm actions at 30Hz from the trained neural networks, and follow the temporal ensembling proposed in~\citet{zhao2023learning} to interpolate robot actions. Arm and gripper actions are predicted as continuous positions, and a continuous servo loop commands the robot arm to follow these commands. The predicted eye actions are converted to motor velocities and smoothed with EMA before being sent to motor controllers. We use random sampling during eye inference, to lessen the distribution shift between train and test. In total, the inference control loop including all neural network passes runs at 30hz, and we employ temporal ensembling to interpolate the output actions. We use $k=0.05$ for the exponential smoothing coefficient. During test rollouts, we keep the eye paused for 2 seconds before allowing the hand to move to allow the eye to search. We allow an extra 1 second during inference since in practice the physical eye moves slightly slower because of its inertia than the inertia-free simulated eye.

\section{Camera Comparison Details}
Our camera comparisons use the same transformer-architecture as FoRT, except we remove all eye-related tokens and pass in a $2\times$ higher resolution ($448$) image to match the number of image tokens as FoRT. When two images (wrist and exo) are present, to enable the same batch size training we set both exo and wrist images to $360$ resolution. Positional encoding for baseline image tokens is the same as for FoRT; with learnable query tokens assigned the x,y coordinates of the center of the image. When two images are present, we offset the x coordinates of the exo image to be side-by-side with the wrist image, and set the learnable coordinates to the center of the wrist image. For wrist and wrist+exo comparisons, we find that relative actions out-perform absolute due to the wrist's highly local viewpoint, and absolute actions outperform on the exo only comparison.

\section{Additional Related Work}

Simulation environments are critical for modern robotics research, where physical experiments can be costly and difficult to reproduce.
While many environments are focused on physics~\cite{coumans2020pybullet,makoviychuk2021isaac,todorov2012mujoco,zakka2025mujoco}, others are tailored to the embodied visual tasks central to active vision.
Habitat renders photorealistic RGB–D scans of indoor scenes, enabling large-scale navigation and manipulation studies~\cite{savva2019habitat,szot2021habitat,puig2023habitat}.  
iGibson couples the Bullet physics engine with textured reconstructions of real homes and offices, offers domain randomization, and exposes interactive objects for embodied learning~\cite{shen2021igibson,li2021igibson}.  
AI2-THOR provides Unity-based apartments with rich object affordances~\cite{kolve2017ai2}; its extensions add real-world counterparts for sim-to-real transfer~\cite{deitke2020robothor}, a tabletop manipulator arm~\cite{ehsani2021manipulathor}, and large procedurally generated layouts~\cite{deitke2022procthor}.
We introduce EyeGym as an alternative environment for training physical eyes to look around in any 360 direction.
Rather than rendering 3D reconstructions, EyeGym builds on the insight that camera observations can be simulated by sampling from any \(360^{\circ}\) image or video~\cite{hu2017deep,wallingford2024image}.
This approach has two key advantages: (i) it minimizes the sim-to-real gap by exposing agents to real-world textures, lighting, and noise, and (ii) it enables large-scale pretraining using native \(360^{\circ}\) video datasets such as \textsc{360-1M}~\cite{wallingford2024image}, which contains one million panoramic YouTube videos with unconstrained egocentric camera motion.  
These properties make EyeGym practical and scalable for learning active visual perception policies that transfer to real-world camera systems.

\end{document}